\pdfoutput=1

\documentclass{llncs}
\usepackage{makeidx}  
\usepackage{color}
\usepackage{graphicx}
\usepackage{amsmath}
\usepackage{amssymb}
\usepackage{etoolbox}

\newtoggle{anon}

\togglefalse{anon}

\def\eg{{\em e.g.,}}

\def\vs{{\em vs. }}
\newcommand{\R}{\mathcal{R}}

\begin{document}
\frontmatter          
\pagestyle{headings}  

\mainmatter              
\title{Detecting Cancer Metastases on\\Gigapixel Pathology Images}
\titlerunning{Detecting Cancer Metastases on Gigapixel Pathology Images}  
%

\iftoggle{anon}{ 

\author{* \thanks{\small *} \thanks{\small *} \\
* \\
* \\
* \\ [.1cm]
\email{\{*@*.*\}} \\ [.2cm]
* \\
*
} 
\authorrunning{*} 
\institute{\vspace*{-.6cm}}

\tocauthor{Yun Liu, Krishna Gadepalli, Mohammad Norouzi, George Dahl, Timo Kohlberger, Aleksey Boyko, Subhashini Venugopalan, Aleksei Timofeev, Philip Nelson, Greg Corrado, Jason Hipp, Lily Peng, Martin Stumpe}

}{
\author{Yun~Liu$^1$\thanks{\small Work done as a Google Brain Resident (g.co/brainresidency).} \and Krishna~Gadepalli$^1$ \and Mohammad~Norouzi$^1$ \and George~E.~Dahl$^1$ \and
Timo~Kohlberger$^1$ \and Aleksey~Boyko$^1$ \and Subhashini~Venugopalan$^2$\thanks{\small Work done as a Google intern.} \and
Aleksei~Timofeev$^2$ \and Philip~Q.~Nelson$^2$ \and Greg~S.~Corrado$^1$ \and Jason~D.~Hipp$^3$ \and 
 Lily~Peng$^1$ \and Martin~C.~Stumpe$^1$\\[.1cm]
\email{\{liuyun,mnorouzi,gdahl,lhpeng,mstumpe\}@google.com}\\[.2cm]
$^1$Google Brain,~$^2$Google Inc,~$^3$Verily Life Sciences,\\
Mountain View, CA, USA
}
\authorrunning{Liu et al.} 
\institute{\vspace*{-.6cm}}

\tocauthor{Yun Liu, Krishna Gadepalli, Mohammad Norouzi, George Dahl, Timo Kohlberger, Aleksey Boyko, Subhashini Venugopalan, Aleksei Timofeev, Philip Nelson, Greg Corrado, Jason Hipp, Lily Peng, Martin Stumpe}
}

\maketitle              

\begin{abstract}

Each year, the treatment decisions for more than $230,000$ breast cancer patients in the U.S. hinge on
whether the cancer has metastasized away from the breast. Metastasis
detection is currently performed by pathologists reviewing large
expanses of biological tissues. This process is labor intensive and
error-prone. We present a framework to automatically detect and localize
tumors as small as $100\!\times\!100$ pixels in gigapixel microscopy
images sized $100,000\!\times\!100,000$ pixels. Our method leverages
a convolutional neural network (CNN) architecture and obtains
state-of-the-art results on the Camelyon16 dataset in the challenging
lesion-level tumor detection task. At $8$ false positives per image, we
detect $92.4\%$ of the tumors, relative to $82.7\%$ by the previous
best automated approach. For comparison, a human
pathologist attempting exhaustive search achieved $73.2\%$ sensitivity.
We achieve image-level AUC scores above $97\%$ on both the Camelyon16 test set
and an independent set of 110 slides.
In addition, we discover that two slides in the Camelyon16 training set were
erroneously labeled normal. Our approach could considerably reduce false
negative rates in metastasis detection.

\keywords{neural network, pathology, cancer, deep learning}

\end{abstract}

\section{Introduction}

The treatment and management of breast cancer is determined by the disease stage. A central component of breast cancer staging involves the microscopic examination of lymph nodes adjacent to the breast for evidence that the cancer has spread, or metastasized \cite{apple2016sentinel}. This process requires highly skilled pathologists and is fairly time-consuming and error-prone, particularly for lymph nodes with either no or small tumors. Computer assisted detection of lymph node metastasis could increase the sensitivity, speed, and consistency of metastasis detection \cite{litjens2016deep}.

In recent years, deep CNNs have significantly improved accuracy on a wide range of computer vision tasks such as image recognition \cite{krizhevsky2012imagenet,ioffe2015batch,russakovsky2015imagenet}, object detection \cite{rcnn}, and semantic segmentation \cite{long2015fully}. Similarly, deep CNNs have been applied productively to improve healthcare (\eg \cite{gulshan2016development}).

This paper presents a CNN framework to aid breast cancer metastasis detection in lymph nodes. We build on \cite{wang2016deep} by leveraging a more recent Inception architecture \cite{szegedy2015going}, careful image patch sampling and data augmentations. Despite performing inference with stride 128 (instead of 4), we halve the error rate at 8 false positives (FPs) per slide, setting a new state-of-the-art. We also found that several approaches yielded no benefits: \textbf{(1)} a multi-scale approach that mimics the human cognition of a pathologist's examination of biological tissue, \textbf{(2)} pre-training the model on ImageNet image recognition, and \textbf{(3)} color normalization. Finally, we dispense with the random forest classifier and feature engineering used in \cite{wang2016deep} and find that the maximum function is an effective whole-slide classification procedure.

\textbf{Related Work}
Several promising studies have applied deep learning to histopathology. The Camelyon16 challenge winner \cite{camelyon2016results} achieved a sensitivity of 75\% at 8 FP per slide and a slide-level classification AUC of 92.5\%~\cite{wang2016deep}. The authors trained a Inception (V1, GoogLeNet) \cite{szegedy2015going} model on a pre-sampled set of image patches, and trained a random forest classifier on 28 hand-engineered features to predict the slide label. A second Inception model was trained on harder examples, and predicted points were generated using the average of the two models' predictions. This team later improved these metrics to 82.7\% and 99.4\% respectively \cite{camelyon2016results} using color normalization \cite{bejnordi2016stain}, additional data augmentation, and lowering the inference stride from 64 to 4. The Camelyon organizers also trained CNNs on smaller datasets to detect breast cancer in lymph nodes and prostate cancer biopsies \cite{litjens2016deep}. \cite{janowczyk2016deep} applied CNNs to segmenting or detecting nuclei, epithelium, tubules, lymphocytes, mitosis, invasive ductal carcinoma and lymphoma. \cite{cruz2014automatic} demonstrated that CNNs achieved higher F1 score and balanced accuracy in detecting invasive ductal carcinoma. CNNs were also used to detect mitosis, winning the ICPR12 \cite{cirecsan2013mitosis} and AMIDA13 \cite{veta2015assessment} mitosis detection competitions. Other efforts at leveraging machine learning for predictions in cancer pathology include predicting prognosis in non-small cell lung cancer \cite{yu2016predicting}.

\section{Methods}

\begin{figure}[!t]
\resizebox{\textwidth}{!}{\begin{tabular}{ccccccc}
\includegraphics{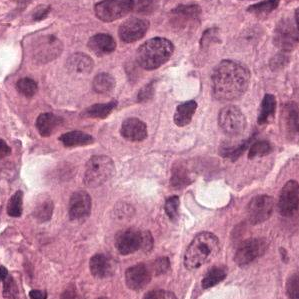}~ & \includegraphics{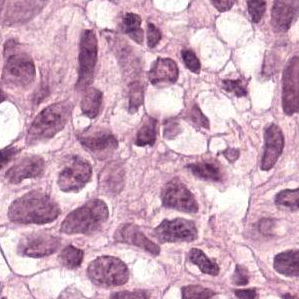}~ & \includegraphics{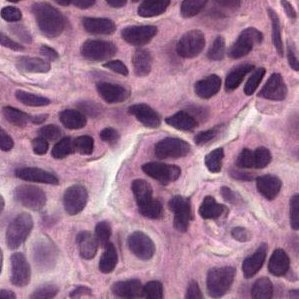} &~~~~~~~~~~& \includegraphics{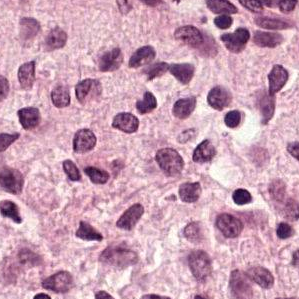}~ & \includegraphics{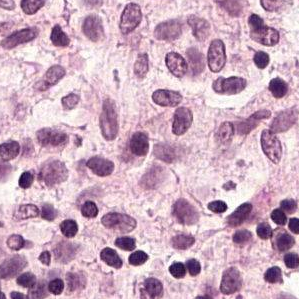}~ & \includegraphics{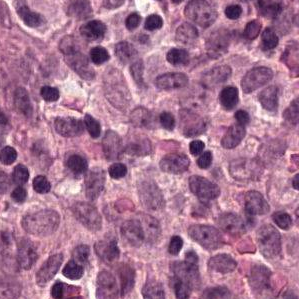} \\
\end{tabular}}
\caption{\textbf{Left}: three tumor patches and \textbf{right}: three challenging normal patches.}
\label{fig:sample_patches}
\end{figure}

\begin{figure}[!t]
    \centering
    \begin{minipage}{0.40\textwidth}
        \centering
        \includegraphics[width=1.0\textwidth]{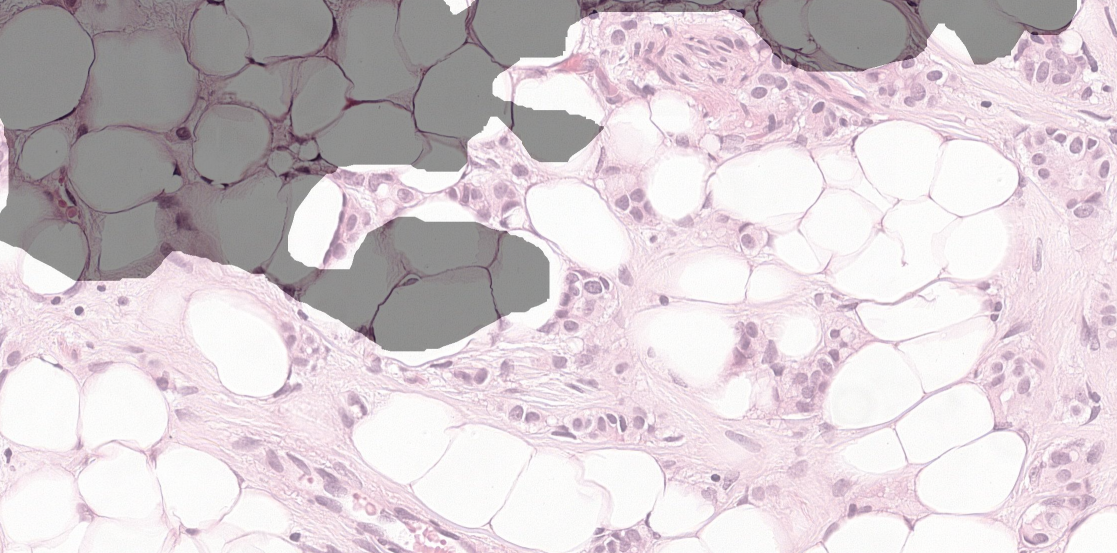}
        \caption{Difficulty of pixel-accurate annotations for scattered tumor cells. Ground truth annotation is overlaid with a lighter shade. Note that the tumor annotations include both tumor cells and normal cells \eg white space representing adipose tissue (fat). \label{fig:inexact_annotation}}
    \end{minipage}\hfill
    \begin{minipage}{0.55\textwidth}
        \centering
        \includegraphics[width=1.0\textwidth]{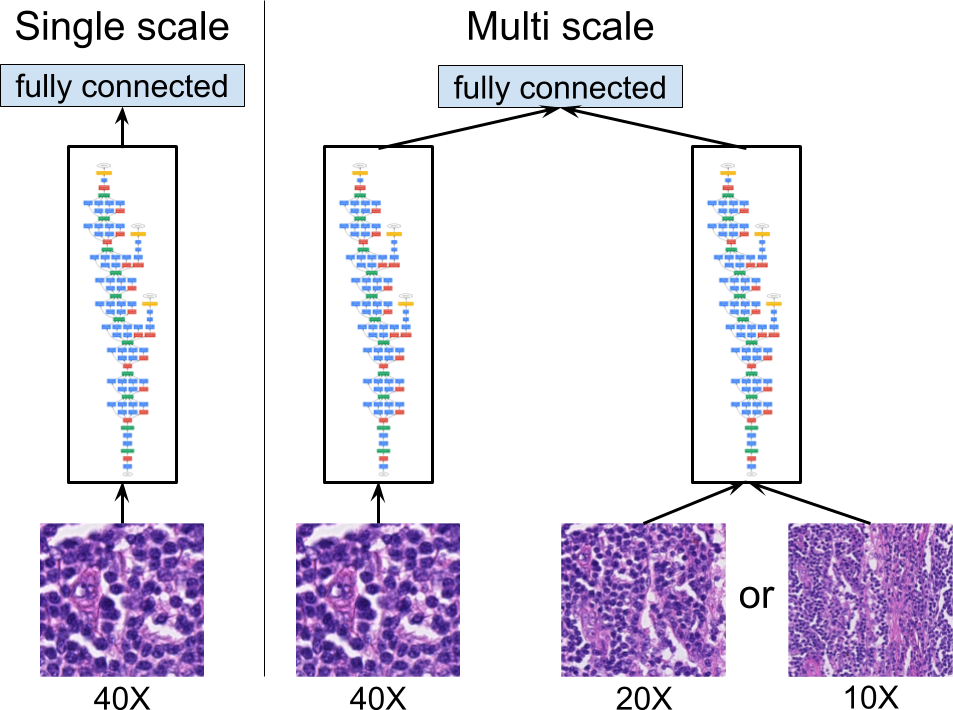}
        \caption{The three colorful blocks represent Inception (V3) towers up to the second-last layer (PreLogit). \emph{Single scale} utilizes one tower with input images at 40X magnification; \emph{multi-scale} utilizes multiple (\eg 2) input magnifications that are input to separate towers and merged.\label{fig:architecture}}
    \end{minipage}
\end{figure}

Given a gigapixel pathology image (\emph{slide}\footnote{\small Each slide contains human lymph node tissue stained with hematoxylin and eosin (H\&E), and is scanned at the most common high magnification in a microscope, ``40X''. We also experimented with 2- and 4-times down-sampled patches (``20X'' and ``10X'').}), the goal is to classify if the image contains tumor and localize the tumors for a pathologist's review. This use case and the difficulty of pixel-accurate annotation (Fig.~\ref{fig:inexact_annotation}) renders detection and localization more important than pixel-level segmentation. Because of the large size of the slide and the limited number of slides ($270$), we train models using smaller image \emph{patches} extracted from the slide (Fig.~\ref{fig:sample_patches}). Similarly, we perform inference over patches in a sliding window across the slide, generating a tumor probability \emph{heatmap}. For each slide, we report the maximum value in the heatmap as the slide-level tumor prediction.

We utilize the Inception (V3) architecture \cite{szegedy2015going} with inputs sized $299\!\times\!299$ (the default) to assess the value of initializing from existing models pre-trained on another domain. For each input patch, we predict the label of the center $128\!\times\!128$ region. A 128 pixel region can span several tumor cells and was also used in \cite{litjens2016deep}. We label a patch as tumor if at least one pixel in the center region is annotated as tumor. We explored the influence of the number of parameters by reducing the number of filters per layer while keeping the number of layers constant (\eg~$depth\_multiplier = 0.1$ in TensorFlow). We denote these models ``small''. We also experimented with multi-scale approaches that utilize patches at multiple magnifications centered on the same region (Fig.~\ref{fig:architecture}). Because preliminary experiments did not show a benefit from using up to four magnifications, we present results only for up to two magnifications.

Training and evaluating our models was challenging because of the large number of patches and the tumor class imbalance. Each slide contains $10,000$ to $400,000$ patches (median $90,000$). However, each tumor slide contains $20$ to $150,000$ tumors patches (median $2,000$), corresponding to tumor patch percentages ranging from $0.01\%$ to $70\%$ (median $2\%$). Avoiding biases towards slides containing more patches (both normal and tumor) required careful sampling. First, we select ``normal'' or ``tumor'' with equal probability. Next, we select a slide that contains that class of patches uniformly at random, and sample patches from that slide. By contrast, some existing methods pre-sample a set of patches from each slide \cite{wang2016deep}, which limits the breadth of patches seen during training.

To combat the rarity of tumor patches, we apply several data augmentations. First, we rotate the input patch by 4 multiples of 90$^\circ$, apply a left-right flip and repeat the rotations. All 8 orientations are valid because pathology slides do not have canonical orientations. Next, we use TensorFlow's image library (\emph{tensorflow.image.random\_X}) to perturb color: brightness with a maximum delta of $64/255$, saturation with a maximum delta of $0.25$, hue with a maximum delta of $0.04$, and contrast with a maximum delta of $0.75$. Lastly, we add jitter to the patch extraction process such that each patch has a small x,y offset of up to $8$ pixels. The magnitudes of the color perturbations and jitter were lightly tuned using our validation set. Pixel values are clipped to $[0,1]$ and scaled to $[-1,1]$.

We run inference across the slide in a sliding window with a stride of 128 to match the center region's size. For each patch, we apply the rotations and left-right flip to obtain predictions for each of the 8 orientations, and average the 8 predictions.

\textbf{Implementation Details}
We trained our networks with stochastic gradient descent in TensorFlow \cite{tensorflow2015-whitepaper}, with 8 replicas each running on a NVIDIA Pascal GPU with asynchronous gradient updates and batch size of 32 per replica. We used RMSProp \cite{tieleman2012lecture} with momentum of 0.9, decay of 0.9 and $\epsilon = 1.0$. The initial learning rate was 0.05, with a decay of 0.5 every 2 million examples. For refining a model pretrained on ImageNet, we used an initial learning rate of 0.002.

\section{Evaluation and Datasets}

We use the two Camelyon16 evaluation metrics \cite{camelyon2016results}. The first metric, the area under receiver operating characteristic, (Area Under ROC, AUC) \cite{hanley1982meaning} evaluates \emph{slide-level} classification. This metric is challenging because of the potential for FPs when $10^5$ patch-level predictions are obtained per slide. We obtained 95\% confidence intervals using a bootstrap approach\footnote{\small Sample with replacement $n$ slides from the dataset/split, where $n$ is the number of slides in the dataset/split, and compute the AUC. Repeat for a total of 2000 bootstrap samples, and report the $2.5$ and $97.5$ percentile values.}.

The second metric, FROC \cite{bunch1977free}, evaluates \emph{tumor detection and localization}. We first generate a list of coordinates and corresponding predictions from each heatmap. Among all coordinates that fall within each annotated tumor region, the highest prediction is retained. Coordinates falling outside tumor regions are FPs. We use these values to compute the ROC. The FROC is defined as the sensitivity at $0.25, 0.5, 1, 2, 4, 8$ average FPs per tumor-negative slide \cite{litjens2016deep}. This metric is challenging because reporting multiple points per FP region can quickly erode the score. We focused on the FROC as opposed to the AUC because there are approximately twice as many tumors as slides, which improves the reliability of the evaluation metric. Similar to the AUC, we report 95\% confidence intervals by computing the FROC over 2000 bootstrap samples of the predicted points. In addition, we report the sensitivity at 8 FP per slide (``@8FP'') to assess the false negative rate.

To generate points for FROC computation, the Camelyon winners \cite{wang2016deep,camelyon2016results} thresholded the heatmap to produce a bit-mask, and reported a single prediction for each connected component in the bit-mask. By contrast, we use a non-maxima suppression method similar to \cite{cirecsan2013mitosis} that repeats two steps until no values in the heatmap remain above a threshold $t$: \textbf{(1)} report the maximum and corresponding coordinate, and \textbf{(2)} set all values within a radius $r$ of the maximum to 0. Because we apply this procedure to the heatmap, $r$ has units of 128 pixels. $t$ controls the number of points reported and has no effect on the FROC unless the curve plateaus before 8 FP. To avoid erroneously dropping tumor predictions, we used a conservative threshold of $t=0.5$.

\textbf{\underline{Datasets}}
Our work utilizes the Camelyon16 dataset \cite{camelyon2016results}, which contains 400 slides: 270 slides with pixel-level annotations, and 130 unlabeled slides as a test set.\footnote{\small The test slides labels were released recently as part of the training dataset for Camelyon17. We used these labels for evaluation, but not for parameter tuning.} We split the 270 slides into train and validation sets (Appendix) for hyperparameter tuning. Typically only a small portion of a slide contains biological tissue of interest, with background and fat comprising the remainder (\eg~Fig.~\ref{fig:inexact_annotation}). To reduce computation, we removed background patches (gray value $>0.8$ \cite{janowczyk2016deep}), and verified visually that lymph node tissue was not discarded.

\textbf{Additional Evaluation: NHO-1}
We digitized another set of 110 slides (57 containing tumor) from H\&E-stained lymph nodes extracted from 20 patients (86 biological tissue blocks\footnote{\small A tissue block can contain multiple slides that vary considerably at the pixel level.}) as an additional evaluation set. These slides came with patient- or block-level labels. To determine the slide labels, a board-certified pathologist blinded to the predictions adjudicated any differences, and briefly reviewed all 110 slides.

\section{Experiments \& Results}

\begin{table}
\begin{center}
\resizebox{\textwidth}{!}{\begin{tabular}{|l|c|c|c|c|c|c|}
\hline
Input \& & \multicolumn{3}{c|}{Validation} & \multicolumn{3}{c|}{Test} \\

model size     & FROC & @8FP & AUC & FROC & @8FP  & AUC \\
\hline
40X            & 98.1 & 100 & 99.0 & \textbf{87.3} (83.2, 91.1) & \textbf{91.1} (87.2, 94.5) & \textbf{96.7} (92.6, 99.6) \\
40X-pretrained & 99.3 & 100 & 100  & \textbf{85.5} (81.0, 89.5) & \textbf{91.1} (86.8, 94.6) & \textbf{97.5} (93.8, 99.8) \\
40X-small      & 99.3 & 100 & 100  & \textbf{86.4} (82.2, 90.4) & \textbf{92.4} (88.8, 95.7) & \textbf{97.1} (93.2, 99.8) \\
ensemble-of-3 & -    & -   & -    & \textbf{88.5} (84.3, 92.2) & \textbf{92.4} (88.7, 95.6) & \textbf{97.7} (93.0, 100) \\
\hline
20X-small      & 94.7 & 100  & 99.6 & \textbf{85.5} (81.0, 89.7) & \textbf{91.1} (86.9, 94.8) & \textbf{98.6} (96.7, 100) \\
10X-small      & 88.7 & 97.2 & 97.7 & 79.3 (74.2, 84.1) & 84.9 (80.0, 89.4) & \textbf{96.5} (91.9, 99.7) \\
40X+20X-small  & 94.9 & 98.6 & 99.0 & \textbf{85.9} (81.6, 89.9) & \textbf{92.9} (89.3, 96.1) & \textbf{97.0} (93.1, 99.9) \\
40X+10X-small  & 93.8 & 98.6 & 100  & 82.2 (77.0, 86.7) & 87.6 (83.2, 91.7) & \textbf{98.6} (96.2, 99.9) \\
\hline
Pathologist \cite{camelyon2016results} & -  & - & - & 73.3* & 73.3* & \textbf{96.6} \\
Camelyon16 winner \cite{camelyon2016results,wang2016deep} & -  & - & - & 80.7 & 82.7 & \textbf{99.4} \\
\hline
\end{tabular}}
\end{center}
\caption{Results on Camelyon16 dataset (95\% confidence intervals, CI). Bold indicates results within the CI of the best model. ``Small'' models contain 300K parameters per Inception tower instead of 20M. -: not reported. *A pathologist achieved this sensitivity (with no FP) using 30 hours.}
\label{table:main_result}
\end{table}

To perform slide-level classification, the current state-of-the-art methods apply a random forest to features extracted from a heatmap prediction \cite{camelyon2016results}. Unfortunately, we were unable to train slide-level classifiers because the 100\% validation-set AUC (Table~\ref{table:main_result}) rendered internal evaluation of improvements impossible. Nonetheless, using the maximum value of each slide's heatmap achieved AUCs $>97\%$, statistically indistinguishable from the current best results.

For tumor-level classification, we find that the connected component approach \cite{wang2016deep} provides a $1-5\%$ gain in FROC when the FROC is modest ($<80\%$), by masking FP regions. However, this approach is sensitive to the threshold (up to $10-20\%$ variance), and can confound evaluation of model improvements by grouping multiple nearby tumors as one. By contrast, our non-maxima suppression approach is relatively insensitive to $r$ between $4$ and $6$, although less accurate models benefited from tuning $r$ using the validation set (\eg~$8$). Finally, we achieve 100\% FROC on larger tumors (macrometastasis), indicating that most false negatives are comprised of smaller tumors.

Previous work (\eg~\cite{yosinski2014transferable,gulshan2016development}) has shown that \emph{pre-training} on a different domain improves performance. However, we find that although pre-training significantly improved convergence speed, it did not improve the FROC (see Table~\ref{table:main_result}: 40X \vs 40X-pretrained). This may be due to a large domain difference between pathology images and natural scenes in ImageNet, leading to limited transferability. In addition, our large dataset size ($10^7$ patches) and data augmentation may have enabled the training of accurate models without pre-training.

Next, we studied the effect of \emph{model size}. Although we were originally motivated by improved experiment turn-around time, we surprisingly found that slimmed-down Inception architectures with only 3\% of the parameters achieved similar performance to the full version (Table~\ref{table:main_result}: 40X \vs 40X-small). Thus, we performed the remaining experiments using this smaller model.

\begin{figure}[t]
\begin{center}
\includegraphics[width=1.0\linewidth]{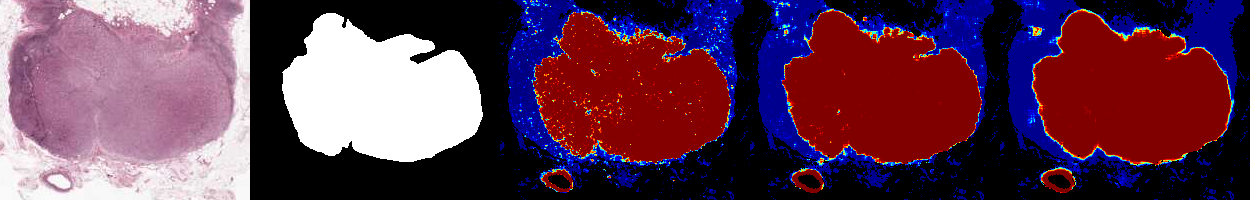}
\end{center}
\caption{Left to right: sample image, ground truth (tumor in white), and heatmap outputs (40X-ensemble-of-3, 40X+20X, and 40X+10X). Heatmaps of 40X and 40X-ensemble-of-3 look identical. The red circular regions at the bottom left quadrant of the heatmaps are unannotated tumor. Some of the speckles are either out of focus patches on the image or non-tumor patches within a large tumor.}
\label{fig:heatmap_multiscale}
\end{figure}

We also experimented with a \emph{multi-scale} approach inspired by pathologists' workflow of examining a slide at multiple magnifications to get context. However, we find no performance benefit in combining 40X with an additional input at lower magnification (Fig.~\ref{fig:architecture}). However, these combinations output smoother heatmaps (Fig.~\ref{fig:heatmap_multiscale}), likely because of translational invariance of the CNN and overlap in adjacent patches. These visual improvements can be deceptive: some of the speckles in the 40X models reveal small non-tumor regions surrounded by tumor.

Figures~\ref{fig:sample_patches} and \ref{fig:architecture} highlight the variability in the images. Although the current leading approaches report improvements from \emph{color normalization}, our experiments revealed no benefit (Appendix). This could be explained by our extensive data augmentations causing our models to learn color-invariant features.

Finally, we experimented with \emph{ensembling} models in two ways. First, averaging predictions across the 8 rotations/flips yielded a few percent improvement in the metrics. Second, ensembling across independently trained models yield additional but smaller improvements, and gave diminishing returns after 3 models.

\textbf{\underline{Additional Validation}}
We also tested our models on another 110 slides that were digitized on different scanners, from different patients, and treated with different tissue preparation protocols. Encouragingly, we obtained an AUC of \textbf{97.6 (93.6, 100)}, on-par with our Camelyon16 test set performance.

\textbf{Qualitative Evaluation}
We discovered tumors in two ``normal'' slides: 086 and 144. Fortunately, the challenge organizers confirmed that both were data processing errors, and the patients were unaffected. Remarkably, both slides were in our training set, suggesting that our model was relatively resilient to label noise. In addition, we discovered an additional 7 tumor slides with incomplete annotations: 5 in train, 2 in validation (Appendix). Samples of our predictions and corresponding patches are shown in the Appendix.

\textbf{Limitations}
Our errors were related to out-of-focus tissues (macrophages, germinal centers, stroma), and tissue preparation artifacts. These errors could be reduced by better scanning quality, tissue preparation, and more comprehensive labels for different tissue types. In addition, we were unable to exhaustively tune our hyperparameters owing to the near-perfect FROC and AUC on our validation set. We plan to further develop our work on larger datasets.

\section{Conclusion}

Our method yields state-of-the-art sensitivity on the challenging task of detecting small tumors in gigapixel pathology slides, reducing the false negative rate to a quarter of a pathologist and less than half of the previous best result. We further achieve pathologist-level slide-level AUCs in two independent test sets. Our method could improve accuracy and consistency of evaluating breast cancer cases, and potentially improve patient outcomes. Future work will focus on improvements utilizing larger datasets.

%
%
%
\bibliographystyle{splncs03}
\pagebreak
\bibliography{references}

\begin{thebibliography}{10}
\providecommand{\url}[1]{\texttt{#1}}
\providecommand{\urlprefix}{URL }

\bibitem{camelyon2016results}
Camelyon 2016. \url{https://camelyon16.grand-challenge.org/}, accessed:
  2017-01-17

\bibitem{tensorflow2015-whitepaper}
Abadi, M., et~al.: {TensorFlow} (2015)

\bibitem{apple2016sentinel}
Apple, S.K.: Sentinel lymph node in breast cancer: Review article from a
  pathologist’s point of view. J. of Pathol. and Transl. Medicine  50(2), ~83
  (2016)

\bibitem{bejnordi2016stain}
Bejnordi, B.E., et~al.: Stain specific standardization of whole-slide
  histopathological images. IEEE Trans. on Medical Imaging  35(2),  404--415
  (2016)

\bibitem{bunch1977free}
Bunch, P.C., et~al.: A free response approach to the measurement and
  characterization of radiographic observer performance. Appl. of Opt. Instrum.
  in Medicine VI pp. 124--135 (1977)

\bibitem{cirecsan2013mitosis}
Cire{\c{s}}an, D.C., et~al.: Mitosis detection in breast cancer histology
  images with deep neural networks. Int. Conf. on Medical Image Comput. and
  Comput. Interv.  (2013)

\bibitem{cruz2014automatic}
Cruz-Roa, A., et~al.: Automatic detection of invasive ductal carcinoma in whole
  slide images with convolutional neural networks. SPIE medical imaging  (2014)

\bibitem{rcnn}
Girshick, R., et~al.: Rich feature hierarchies for accurate object detection
  and semantic segmentation. In: Comput. Vis. and Pattern Recognit. (2014)

\bibitem{gulshan2016development}
Gulshan, V., et~al.: Development and validation of a deep learning algorithm
  for detection of diabetic retinopathy in retinal fundus photographs. J. of
  the Am. Medical Soc.  316(22),  2402--2410 (2016)

\bibitem{hanley1982meaning}
Hanley, J.A., McNeil, B.J.: The meaning and use of the area under a receiver
  operating characteristic (roc) curve. Radiology  143(1),  29--36 (1982)

\bibitem{ioffe2015batch}
Ioffe, S., Szegedy, C.: Batch normalization: Accelerating deep network training
  by reducing internal covariate shift. Int. Conf. on Machine Learning  (2015)

\bibitem{janowczyk2016deep}
Janowczyk, A., Madabhushi, A.: Deep learning for digital pathology image
  analysis: A comprehensive tutorial with selected use cases. J. of Pathol.
  Informatics  7 (2016)

\bibitem{kothari2013pathology}
Kothari, S., et~al.: Pathology imaging informatics for quantitative analysis of
  whole-slide images. J. of the Am. Medical Informatics Assoc.  20(6),
  1099--1108 (2013)

\bibitem{krizhevsky2012imagenet}
Krizhevsky, A., et~al.: Imagenet classification with deep convolutional neural
  networks. Adv. in Neural Inf. Process. Syst. pp. 1097--1105 (2012)

\bibitem{van2000hue}
van~der Laak, J.A., et~al.: Hue-saturation-density model for stain recognition
  in digital images from transmitted light microscopy. Cytometry  39(4),
  275--284 (2000)

\bibitem{litjens2016deep}
Litjens, G., et~al.: Deep learning as a tool for increased accuracy and
  efficiency of histopathological diagnosis. Sci. Reports  6 (2016)

\bibitem{long2015fully}
Long, J., et~al.: Fully convolutional networks for semantic segmentation (2015)

\bibitem{pitie2007linear}
Piti{\'e}, F., Kokaram, A.: The linear monge-kantorovitch linear colour mapping
  for example-based colour transfer  (2007)

\bibitem{russakovsky2015imagenet}
Russakovsky, O., et~al.: Imagenet large scale visual recognition challenge.
  Int. J. of Comput. Vis.  115(3),  211--252 (2015)

\bibitem{szegedy2015going}
Szegedy, C., et~al.: Going deeper with convolutions. Comput. Vis. and Pattern
  Recognit.  (2015)

\bibitem{tieleman2012lecture}
Tieleman, T., Hinton, G.: Lecture 6.5-rmsprop: Divide the gradient by a running
  average of its recent magnitude  (2012)

\bibitem{veta2015assessment}
Veta, M., et~al.: Assessment of algorithms for mitosis detection in breast
  cancer histopathology images. Medical image analysis  20(1),  237--248 (2015)

\bibitem{wang2016deep}
Wang, D., et~al.: Deep learning for identifying metastatic breast cancer. arXiv
  preprint arXiv:1606.05718  (2016)

\bibitem{yosinski2014transferable}
Yosinski, J., et~al.: How transferable are features in deep neural networks?
  Adv. in Neural Inf. Process. Syst.  (2014)

\bibitem{yu2016predicting}
Yu, K.H., et~al.: Predicting non-small cell lung cancer prognosis by fully
  automated microscopic pathology image features. Nat. Commun.  7 (2016)

\end{thebibliography}


\appendix
\section{Appendix}

\subsection{Dataset Details}

\begin{table}
\begin{center}
\resizebox{\textwidth}{!}{\begin{tabular}{|l|c|c|c|c|c|c|c|c|c|}
\hline
           & \multicolumn{3}{c|}{Number of Slides} & \multicolumn{3}{c|}{Number of Patches (M)} & \multicolumn{2}{c|}{Number of Tumors} \\
Dataset/split       & Normal & Tumor & Total & Normal & Tumor & Total & Macro & Micro \\
\hline
Camelyon-Train      & 127 & 88 & 215 & 13+8.9*  & 0.87 & 23  & 81 & 345 \\
Camelyon-Validation & 32  & 22 & 54  & 3.8+2.3* & 0.28 & 6.4 & 14 & 58  \\
\hline
Camelyon-Test       & 80  & 50 & 130 &          &      &     & 40 & 185 \\
NHO-1 $\star$       & 53  & 57 & 110 &          &      &     &    &     \\
\hline
\end{tabular}}
\end{center}
\caption{Number of slides, patches (in millions), and tumors in each dataset/split. We excluded ``Normal'' slide 144 because preliminary experiments uncovered tumors in this slide. Later experiments also uncovered tumors in ``Normal'' 086, but this slide was used in training for the results presented in this paper. In addition, Test slide 049 was an accidental duplication by the organizers (Tumor 036), and was not used for evaluation. Tumor sizes: macrometastasis (macro, $> 2000 \mu m$), micrometastasis (micro, $> 200~\& \le 2000 \mu m$). *normal patches extracted from the tumor slides. $\star$: additional evaluation set with slide-level labels only.}
\label{table:data_stats}
\end{table}

\subsection{Soft Labels}

Our experiments used binary labels: a patch is positive if at least one pixel in the center 128 x 128 region is annotated as tumor. We also explored an alternative ``soft label'' approach in preliminary experiments, assigning as the label the fraction of tumor pixels in the center region. However, we found that the thresholded labels yielded substantially better performance. Because the FROC rewards detecting tumors of all size equally, this might reflect the model being trained to assign lower values to smaller tumors (where on average, a smaller portion of each patch contains tumor cells).

\subsection{Image Color Normalization}

As can be seen in Fig.~\ref{fig:sample_patches} \& \ref{fig:architecture}, the (H\&E) stained tissue vary significantly in color. These variations arise from differences in the underlying biological tissue, physical and chemical preparation of the slide, and scanner adjustments. Because reducing these variations have improved performances in other automated detection systems \cite{bejnordi2016stain,kothari2013pathology}, we experimented with a similar color normalizing approach. However, we have not found this normalization to improve performance, and thus we detail our approach for reference only. This lack of improvement likely stems from our extensive color perturbations encouraging our models to learn color-insensitive features, and thus the color normalization was unnecessary.

First, we separate color and intensity information by mapping the raw RGB values to a Hue-Saturation-Density (HSD) space \cite{van2000hue}, and then normalize each component separately. This maps each color channel $ (I_R,I_G, I_B) \in [0,255]^3$ to a corresponding optical density value: $D_\nu = -\ln((I_\nu + 1)/257), \nu \in \{R,G,B\}$, followed by applying a common Hue-Saturation-Intensity color space transformation with $D = (D_R + D_B + D_G)/3$ being the intensity value, and $c_x = \frac{D_R}{D}-1$ and $c_y = (D_G-D_B)/(\sqrt{3}\cdot D)$ denoting the Cartesian coordinates that span the two-dimensional hue-saturation plane. We chose the HSD mapping over a direct HSI mapping of RGB values \cite{van2000hue}, because it is more compatible with the image acquisition physics and yields more compact distributions in general.

Next, we fit a single Gaussian to the color coordinates $(c_x, c_y)_i$ of the pixels in all tissue-containing patches, i.e. compute their empirical mean $\mu = (\mu_x, \mu_y)^T$ and covariance $\Sigma \in \R^{2\times 2}$, and then determine the transformation $T\in \R^{2\times 2}$ of the covariance $\Sigma$ to a reference covariance matrix~$\Sigma^R$ using the Monge-Kantorovitch approach presented in \cite{pitie2007linear}: $T = \Sigma^{-1/2} \left( \Sigma^{1/2} \Sigma_R \Sigma^{1/2} \right) \Sigma^{-1/2}$. Subsequently, we normalize the color values by applying the mapping:

\begin{equation}
\label{eq:color_norm}
\begin{bmatrix}
c'_x \\
c'_y
\end{bmatrix}
= 
T
\left(
\begin{bmatrix}
c_x \\
c_y
\end{bmatrix}
-\begin{bmatrix}
\mu_x \\
\mu_y
\end{bmatrix}
\right)
+\begin{bmatrix}
\mu^R_x \\
\mu^R_y
\end{bmatrix} .
\end{equation}

Intensity values, ${D_i}$, are normalized in the same manner, i.e. by applying the one-dimensional version of Equation~\ref{eq:color_norm} in order to transform the empirical mean and variance of all patch intensities to a reference intensity mean and variance.

As reference means and variances for the color and intensity component, respectively (i.e. $\mu^R_v, \Sigma^R$ for color), we chose the component-wise medians over the corresponding statistical moments of all the training slides.

Finally, we map the normalized $(c'_x, c'_y, D')$ values back to RGB space by first applying the inverse HSI transform \cite{van2000hue}, followed by inverting the non-linear mapping, i.e. by applying $I_\nu = \exp(-D_\nu) \cdot 257-1$ to each component~$\nu \in \{R,G,B\}$.

We applied this normalization in two ways. First, we applied this at inference only, by testing a model (``40X-small'' in Table~\ref{table:main_result}) on color-normalized slides. Unfortunately, this resulted in a few percent drop in FROC. Next we trained two models on color-normalized slides, both with and without the color perturbations. We then tested these models on color-normalized slides. Neither approach improved the performance.

\subsection{Sample Results}

\begin{figure}
\centering
\includegraphics[width=\textwidth]{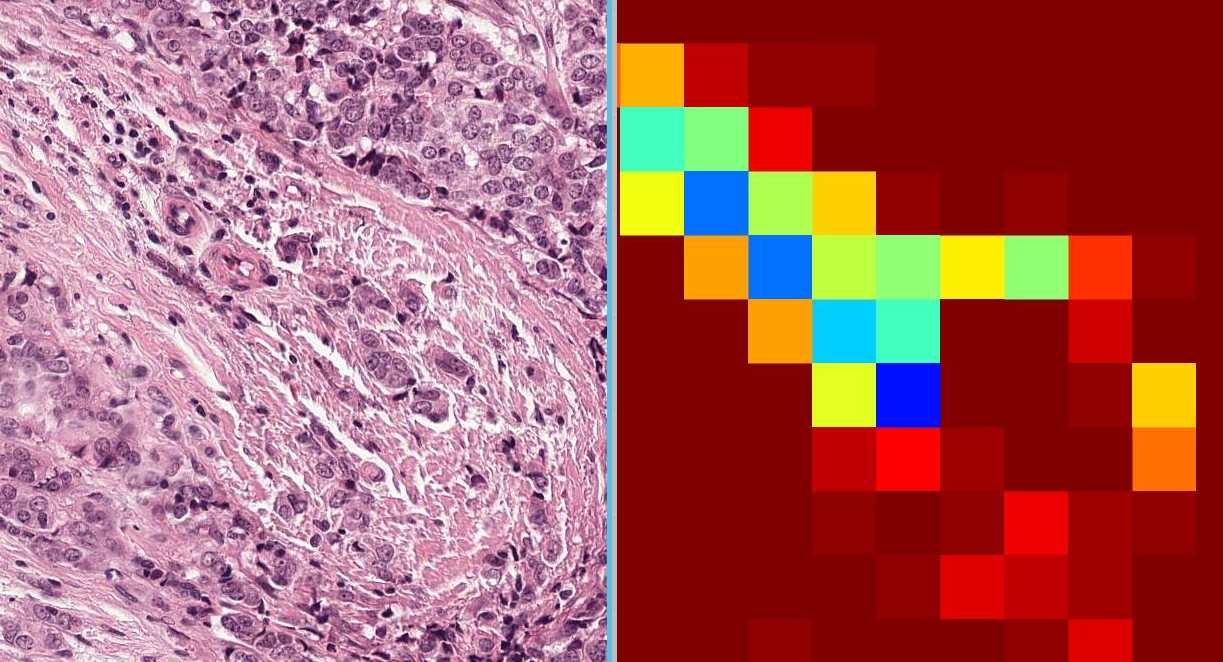}
\caption{\textbf{Left}: a patch from a H\&E-stained slide. The darker regions are tumor, but not the lighter pink regions. \textbf{Right}: the corresponding predicted heatmap that accurately identifies the tumor cells while assigning lower probabilities to the non-tumor regions.}
\label{fig:speckle}
\end{figure}

\begin{figure}
\centering
\includegraphics[width=\textwidth]{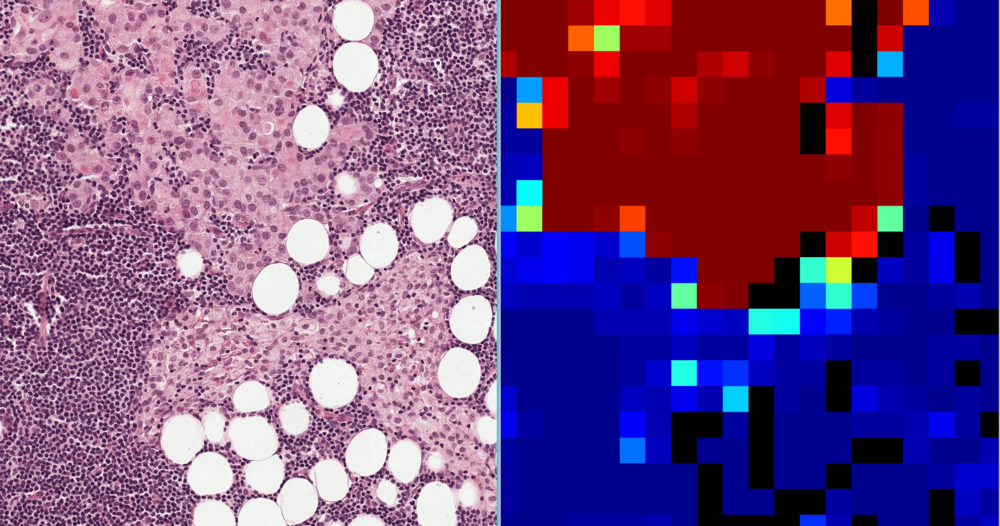}
\caption{\textbf{Left}: a patch from a H\&E-stained slide, ``Normal'' 086. The larger pink cells near the top are tumor, while the smaller pink cells at the bottom are macrophages, a normal cell. \textbf{Right}: the corresponding predicted heatmap that accurately identifies the tumor cells while ignoring the macrophages.}
\label{fig:N086}
\end{figure}

\begin{figure}
\centering
\includegraphics[width=\textwidth]{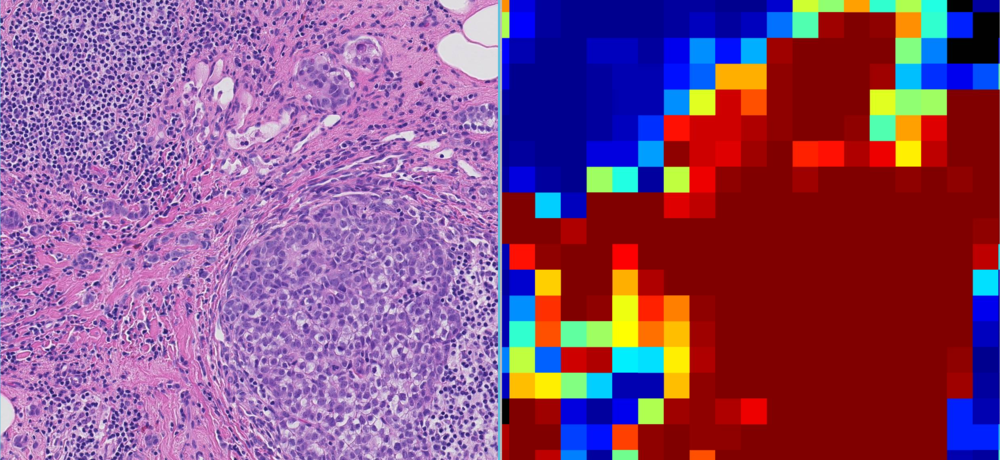}
\caption{\textbf{Left}: a patch from a H\&E-stained slide, ``Normal'' 144. The cluster of larger, dark purple cells in the bottom right quadrant are tumor, while the smaller dark purple cells are lymphocytes. The pink areas are connective tissue, with interspersed tumor cells. \textbf{Right}: the corresponding predicted heatmap that accurately identifies the tumor cells while ignoring the connective tissue and lymphocytes.}
\label{fig:N144}
\end{figure}

\begin{figure}
\centering
\includegraphics[width=\textwidth]{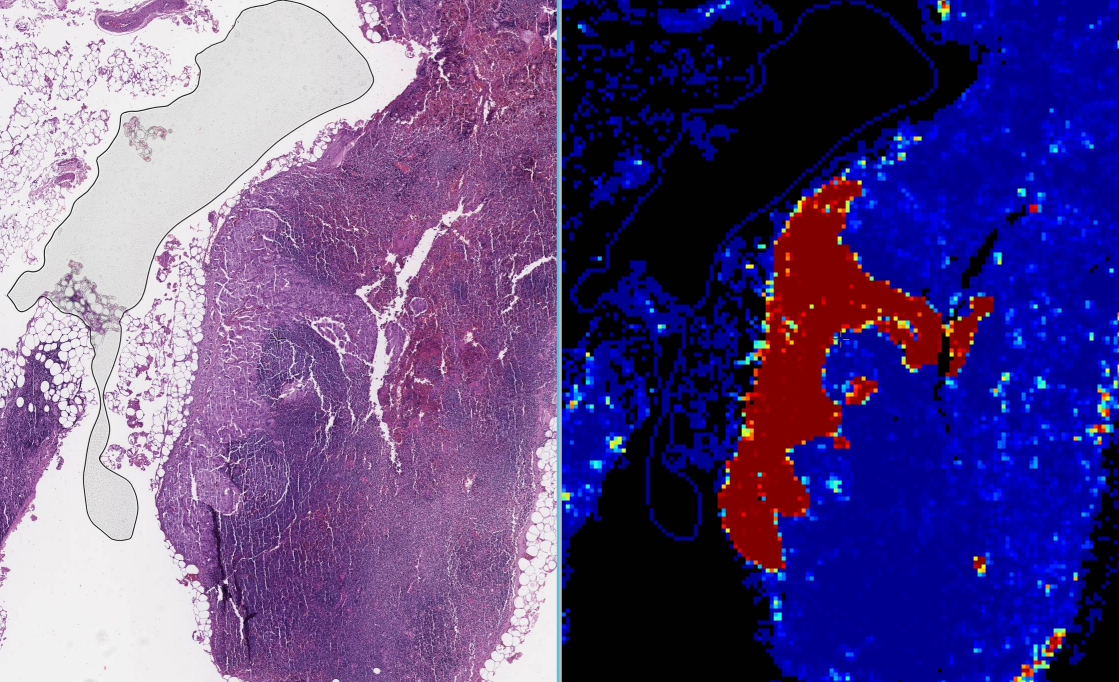}
\caption{\textbf{Left}: a patch from a H\&E-stained slide in our additional validation set, NHO-1. The tumor cells are a lighter purple than the surrounding cells. A variety of artifacts are visible: the dark continuous region in the top left quadrant is an air bubble, and the white parallel streaks in the tumor and adjacent tissue are cutting artifacts. Furthermore, the tissue is hemorrhagic, necrotic and poorly processed, leading to color alterations to the typical pink and purple of a H\&E slide. \textbf{Right}: the corresponding predicted heatmap that accurately identifies the tumor cells while ignoring the various artifacts, including lymphocytes and the cutting artifacts running through the tumor tissue.}
\label{fig:NHO1}
\end{figure}

\textbf{Tumor slides with incomplete annotations} At the outset, 11 tumor slides were known to have non-exhaustive pixel level annotations: 015, 018, 020, 029, 033, 044, 046, 051, 054, 055, 079, 092, and 095. Thus, we did not use non-tumor patches from these slides as training examples of normal patches. Over the course of our experiments, we discovered several more such cases that we verified with a pathologist: 010, 025, 034, 056, 067, 085, 110.

\end{document}